\pdfoutput=1

\documentclass[11pt]{article}

\usepackage[]{acl}

\usepackage{times}
\usepackage{latexsym}

\usepackage[T1]{fontenc}

\usepackage[utf8]{inputenc}

\usepackage{microtype}
\usepackage{amsmath}
\usepackage{amssymb}
\usepackage{mathtools}
\usepackage{amsthm}
\usepackage{amsfonts}
\usepackage{graphicx}
\usepackage{multirow}
\usepackage[normalem]{ulem}
\usepackage{booktabs}
\usepackage{enumitem}
\usepackage{soul}
\usepackage{placeins}
\usepackage{tikz}

\usepackage[capitalize,noabbrev]{cleveref}

\title{Parallel Structures in Pre-training Data Yield In-Context Learning}

\author{Yanda Chen\textsuperscript{1}~~~~~~Chen Zhao\textsuperscript{2,3}~~~~~~Zhou Yu\textsuperscript{1}~~~~~~Kathleen McKeown\textsuperscript{1}~~~~~~He He\textsuperscript{2}\\
\textsuperscript{1}Columbia University,~~\textsuperscript{2}New York University,~~\textsuperscript{3}NYU Shanghai\\\\
{\tt \{yanda.chen, kathy\}@cs.columbia.edu, cz1285@nyu.edu}\\
{\tt zy2461@columbia.edu, hehe@cs.nyu.edu}}

\begin{document}
\maketitle

\begin{abstract}
Pre-trained language models (LMs) are capable of in-context learning (ICL): they can adapt to a task with only a few examples given in the prompt without any parameter update.
However, it is unclear where this capability comes from
as there is a stark distribution shift between pre-training text and ICL prompts.
%
In this work, we study what patterns of the pre-training data contribute to ICL.
We find that LMs' ICL ability depends on 
\textit{parallel structures} in the pre-training data---pairs of phrases following similar templates in the same context window. 
Specifically, we detect parallel structures by checking whether training on one phrase improves prediction of the other, and conduct ablation experiments to study their effect on ICL.
We show that removing parallel structures in the pre-training data reduces LMs' ICL accuracy by {51\%} (vs 2\% from random ablation).
This drop persists even when excluding common patterns such as n-gram repetitions and long-range dependency,
showing the diversity and generality of parallel structures.
A closer look at the detected parallel structures indicates that they cover diverse linguistic tasks and span long distances in the data. 
\end{abstract}

\section{Introduction}
A surprising ability that emerged from language model pre-training is
in-context learning (ICL); ICL allows LMs to adapt to a task given merely a few input-output pairs in the prompt without any parameter update \cite{brown2020language, chowdhery2023palm}.
It is the basis for chain-of-thought  reasoning \cite{wei2022chain} and is widely used to steer model behavior \cite{lin2021truthful,sun2023principle}.
However, it is still unclear how this ability emerges from learning to predict the next word in natural text.
While previous work has shown that transformers can acquire ICL when trained on sequences of in-context examples (i.e.\ concatenations of input-output pairs from a task) \cite{chen2022meta,garg2022what,chan2022data},
real pre-training data is quite different from 
in-context examples.
A better understanding of the source of ICL may help explain other emergent abilities of pre-trained LMs \cite{wei2022emergent,lu2023emergent} and predict when they might fail.

In this work, 
we adopt a data-centric perspective 
and study the question:
\textit{What structures of the pre-training data yield ICL?}
This question is underexplored due to the scale of data and compute required.
As a result, prior work has mainly focused on synthetic data \cite{xie2021explanation}, 
in-context examples \cite{chan2022data}, coarse data properties such as size and domain \cite{shin2022effect}, 
or task-specific data selection
\cite{han-etal-2023-understanding}.

\begin{figure}[t]
\centering
\includegraphics[width=\columnwidth]{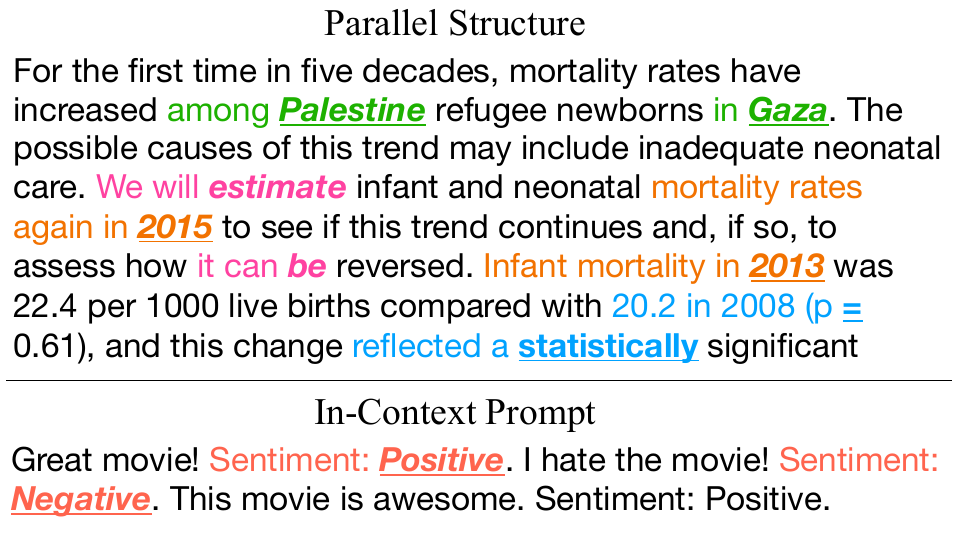}
\caption{\label{fig1} Parallel structures vs. In-context prompts. We define a parallel structure (PS) as two phrases in the window that follow the same distribution. Each phrase consists of a context and a token (bold).
While natural language is unlikely to contain abundant in-context prompts, it often contains parallel structures that exhibit diverse semantic (underlined) and syntactic (italic) patterns.
We hypothesize that parallel structures are essential for LMs to acquire ICL (Section~\ref{sec:icl-and-ps}).
}
\end{figure}

We introduce a simple structure that produces ICL and verify it through ablation on real pre-training data.
Our key observation is that 
while natural language is unlikely to contain abundant in-context examples, it often contains multiple phrases following a similar template within a context window
(Figure~\ref{fig1}), e.g., ``\emph{We will estimate infant and neonatal \ul{morality rates again in 2015} to see if this trend \emph{[...]} reversed. \ul{Infant mortality in 2013} was 22.4 per 1000}.''
These phrases can thus be considered as examples from the same ``task'',
resembling in-context examples.
This motivates us to hypothesize that such co-occurring phrases in pre-training data are essential for LMs to acquire ICL (Section~\ref{sec:icl-and-ps}).

To formalize our hypothesis, we introduce the concept of \textit{parallel structure} (PS), defined as a pair of phrases that co-occur in a context window and follow the same distribution.
To detect PSs in the pre-training data,
our algorithm is based on the intuition that,
since the two phrases are sampled from the same distribution, learning to predict one phrase should improve prediction on the other (Figure~\ref{fig2}).
To verify our hypothesis, we measure the effect of PSs on the model's ICL ability.
Specifically, we
ablate the detected PSs,
train an LM on the ablated data,
and measure the ICL performance drop relative to a reference model trained on clean data
(Section~\ref{sec:experiments}). 

Results on GPT-2 model series \cite{radford2019language} and OpenWebText \cite{Gokaslan2019OpenWeb} show that ablating PSs in the pre-training data significantly reduces the 
ICL accuracy of LMs with a relative decrease of \textbf{51\%}, while ablating randomly sampled tokens of the same amount only reduces ICL accuracy by 2\%.
Furthermore, this effect holds as we increase model size.
This result indicates that PSs are a major source of ICL (Section~\ref{sec:results}).
We also compare PSs to two other structures suggested by prior work as sources of ICL: repetitions \cite{yan2023understanding, olsson2022context} and long-range dependency \cite{shi2023context}, and find that PSs have a larger effect on ICL.

By analyzing characteristics of the detected PSs, we find that they are suggestive of 
ICL abilities we observe in large LMs.
For example, parallel structures exhibit diverse pattern matching tasks,
ranging from n-gram repetitions, 
text formats, syntactic constituents,
to more complicated ones that require reasoning and knowledge.
Pre-training on such a huge diversity of tasks may explain why LMs can generalize to various downstream tasks through ICL \cite{raventos2023pretraining}.
In addition, we find that the two phrases in a PS are often far from each other (343 tokens away on average), which may explain why
LMs don't forget early examples in in-context prompts and why ICL performance improves with more examples \cite{li-qiu-2023-finding}.

\section{Problem Statement}


\paragraph{Pre-trained LMs}
Autoregressive LMs are pre-trained on natural text to predict the next token conditioned on the context. The pre-training dataset $D$ consists of a sequence of context windows
$a=(a_1,\ldots,a_L)$, where $a_i$ denotes the $i$-th token in it.
An LM is a distribution over a token given its prefix.
The parameters of this distribution $w$ are typically learned by maximum likelihood estimation:
\begin{align}
    \mathrm{maximize} \sum_{a \in D}\sum_{i=1}^{L}\log p(a_{i} \mid a_{<i}; w) \;.
\end{align}

\paragraph{In-Context Learning (ICL)}
To adapt a pre-trained LM to a task via ICL, it is prompted with {\it in-context examples}, which is the concatenation of
a sequence of input-output examples of the task: $c_1 \circ x_1 \circ \cdots \circ c_k \circ x_k$, where $c_i$ and $x_i$ denote the task input and output,
and $\circ$ denotes concatenation of two strings.
To make predictions, a test input $c_{\text{query}}$ is appended to the in-context examples
to form an {\it in-context prompt},
and the model predicts the output as the next word distribution given the prompt:
$p(\cdot\mid c_1 \circ x_1 \circ \cdots \circ c_k \circ x_k \circ c_{\text{query}})$.

Since there is a clear divergence between the pre-training data distribution (natural text)
and the in-context prompt distribution (concatenations of task input-output pairs),
it is unclear where LMs acquire their ICL ability from pre-training. 
To bridge this gap,
we aim to identify pre-training examples---tokens and their prefixes---that have large impact on the ICL performance of LMs.

\section{Parallel Structures}
\label{sec:icl-and-ps}

While the pre-training data does not contain a large number of strict in-context prompts,
we observe that it often contains  phrases following a similar template in the same context window. 
These phrase pairs resemble in-context examples of a shared ``task'', but they are less structured.
As shown in \cref{fig1},
they cover a diverse range of linguistic skills, including n-gram copying (e.g., ``\textit{mortality rates again in 2015}'' and ``\textit{infant mortality in 2013}''), syntactic construction (e.g., ``\textit{We will estimate}'' and ``\textit{it can be}'' share the template of subject--modal verb--main verb), world knowledge (e.g., ``\textit{among Palestine}'' and ``\textit{in Gaza}'' mention locations in the same geographical region) and so on.


We conjecture that these co-occurring phrases following similar templates,
termed {\it parallel structures},
are critical for LMs to develop ICL ability during pre-training.
In the rest of this section, we first formally define parallel structures (\mbox{Section~\ref{sec:definition}});
we then propose an algorithm to detect them in natural text (Section~\ref{sec:detection});
finally, we describe how to measure their effect on ICL ability of pre-trained LMs through ablation (Section~\ref{sec:ablation}).

\subsection{Definition}
\label{sec:definition}
Intuitively, phrases following the same template are from the same distribution.
A {\it phrase} is a 
sequence of tokens
and we represent each {phrase} as a (context, token) tuple, $(c, x)$, where $x$ is the the last token in the sequence and $c$ is its prefix,
e.g., (``mortality rates again in'', ``2015'').
Given a context window, a \textit{parallel structure} (PS), denoted by $s$, consists of a pair of phrases in the window
that follow the same distribution $p^s_{\text{struct}}(c, x)$.
We use $(c_f, x_f)$ to denote the {\it former phrase}, which occurs before the {\it latter phrase} $(c_l, x_l)$ in the context window.
For example, given the context window ``increase among Palestine refugee newborns in Gaza'', ($c_f$=``among'', $x_f$=``Palestine'') and ($c_l$=``in'', $x_l$=``Gaza'') form a PS, both following a distribution of prepositional phrases for locations in a specific area. 

\subsection{Finding Parallel Structures in Natural text}
\label{sec:detection}

To study the effect of PSs on ICL, a natural solution is to compare the ICL ability after ablating PSs from the pre-training data, which requires us to first detect them.
Toward this goal, we first define a measure to estimate whether two given phrases come from the same distribution (i.e.\ whether they form a PS according to our definition).
Next, we introduce an efficient algorithm to identify PSs approximately from a large dataset of natural text.

\begin{figure}[t]
\centering
\includegraphics[width=\columnwidth]{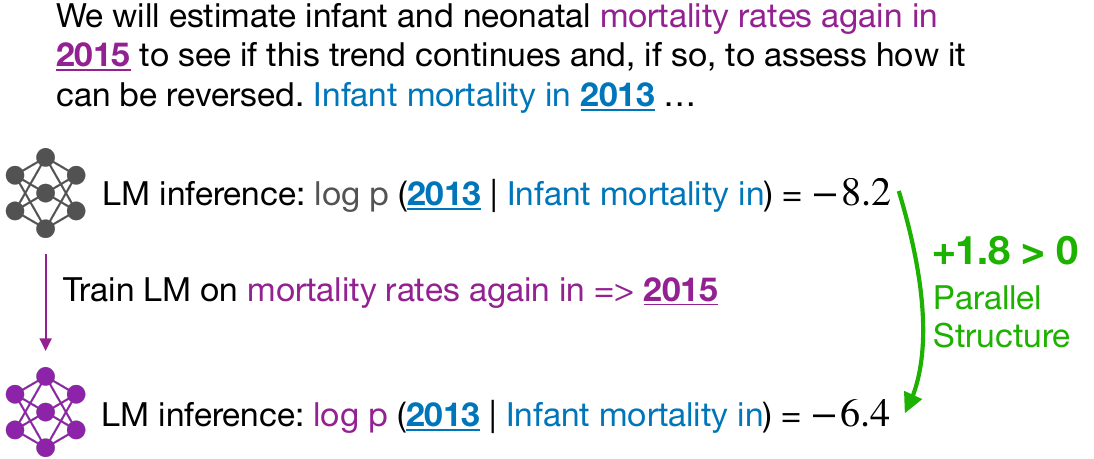}
\caption{\label{fig2} To measure the parallel structure strength of two phrases $(c_f, x_f)$ and $(c_l, x_l)$, we take a pre-trained LM (gray), fine-tune it on $x_f$ conditioned on its context $c_f$ (purple), and measure the change in its predicted probability on $x_l$ conditioned on context $c_l$ (blue). 
}
\end{figure}

\paragraph{Measuring parallel structure strengths.}
Given two phrases, how do we know if they come from the same distribution?
Since we only have two data points, most statistical tests won't apply.
Following the standard supervised learning guarantee with the i.i.d. assumption,
if they  come from the same distribution,
then training on one phrase would improve prediction on the other in general.
In other words, we can think of $(c_f, x_f)$ and $(c_l, x_l)$ as two examples for the  task of predicting $x$ given $c$.
Motivated by this intuition, we measure the {\it parallel structure strength} of two phrases by how much the loss of the latter phrase is reduced from 
training on the former phrase.
A larger reduction suggests better generalization from the former phrase to the latter phrase,
which indicates that they are likely to come from similar distributions.
As shown in Figure~\ref{fig2}, we measure the PS strength of two phrases $(c_f, x_f)$ and $(c_l, x_l)$ by training an LM on the former phrase and test it on the latter.
Formally, given an auto-regressive LM $p(\cdot; w)$ parametrized by $w$, 
we update $w$ 
using the negative log-likelihood loss for one gradient descent step with learning rate $\eta$:
\begin{equation}
    w_{f} = w + \eta \nabla_w \log p(x_f \mid c_f; w) \;.
    \label{eqn:update}
\end{equation}
Then, the PS strength of the phrase pair is measured by the difference  between the log likelihood of the latter token conditioned on its context given by the LM before and after the update:
\begin{align}
    &\alpha((c_f, x_f), (c_l, x_l)) \\
    = &\log p(x_l \mid c_l; w_f) - \log p(x_l \mid c_l; w) \;,
\end{align}
where $\alpha \in \mathbb{R}$ and larger $\alpha$ means stronger PS strength.

\paragraph{Detection algorithm.}

Given a context window (i.e.\ a sequence of tokens) from the pre-training data, $a=(a_1,\ldots,a_L)$,
our goal is to score the PS strength of all pairs of phrases in it and take the top ones as the identified PSs.
However, the naive scoring strategy that enumerates all spans in the window has 
quadratic complexity in the window size $L$,
and is prohibitively expensive when scaled to the pre-training dataset.
Therefore, we apply two approximations for efficiency.
First, we only score a subset of phrase pairs.
Second, we train the LM on a group of former phrases instead of training on each one separately as in \cref{eqn:update}.
We describe the process in detail below.

At a high level, to compute the PS strength, we need to come up with a set of former phrases, update the LM on each phrase, and test the LM on the corresponding latter phrases.
To come up with the former and latter phrases, 
we first decide the last token in a phrase;
then, instead of enumerating prefixes of varying lengths,
we set the prefix of former phrases to be all tokens before the last token, and the prefix of latter phrases to be all tokens before the last token 
limited in a segment of the context window.
We set the prefix of latter phrases to be short to pinpoint the exact latter phrase that forms a PS with preceding tokens in the context window, which we will then ablate.
Specifically, given a context window $a$, we create a set of former phrases $D_f(a) = \{(a_{<i}, a_i)\}_{i=2}^{L}$.
To create the set of latter phrases, we partition the context window $a$ into overlapping segments of length $m$ with stride $m/2$. 
Let $B$ be the set of all such segments in $a$.
We then extract latter phrases from each segment:
$D_l(a) = \bigcup_{b\in B} \{(b_{<i}, b_i)\}_{i=m/2}^m$.
\begin{center}
\begin{tikzpicture}[scale=0.9]
     \fill[red!20] (0,0) rectangle (8,0.5);
     \node[anchor=west] (a) at (0,0.25) {$\color{red}{a}$};
     \node[anchor=west] (xf) at (0+3,0.25) {$x_f$};

     \fill[green!20] (4,0) rectangle ++(2,0.5);
     \node[anchor=west] (b) at (4,0.25) {$\color{green!50!black}{b}$};

     \node[anchor=west] (xl) at (4+1.25,0.25) {$x_l$};
     \draw[-latex, thick, <->] (xf) -- node[above,midway]{$c_f$} (a);
     \draw[-latex, thick, <->] (xl) -- node[above,midway]{$c_l$} (b);
     \draw[-latex, thick, <->, green!50!black] (4, -0.3) -- node[above,midway]{$m$} ++(2, 0);
    \draw[-latex, thick, <->, red] (0, -0.45) -- node[above,pos=0.2]{$L$} ++(8, 0);
 \end{tikzpicture}
\end{center}
%
%
Note that the prefix length of former phrases range from $1$ to $L-1$,
whereas the prefix length of latter phrases range from $m/2$ to $m$, limited by the segment $b$.
Instead of enumerating all phrase pairs, we only consider phrases in $D_f(a)$ and $D_l(a)$.

Now, for each former phrase in $D_f(a)$, we can update an LM $p(\cdot; w)$ on it,
and test the updated LM on latter phrases in $D_l(a)$ that occur after the former phrase (i.e.\ their last tokens occur after the last token of the former phrase).
However, this requires us to perform $\Theta(L)$ independent updates of $p(\cdot; w)$
and the gradient computation cannot be batched (as we need the $w_f$ after each update).
To save compute, we sort the former phrases in $D_f(a)$ by the position of the last token of each phrase and split them into batches of size $l$.
For each former phrase $(c_f, x_f)$ in a batch $B_f$, we approximate the update in \cref{eqn:update} by
a minibatch update on all $l$ phrases in the batch:
\begin{align}
w_f = w + \frac{\eta}{l}\sum_{(c, x)\in B_f} \nabla_w\log p(x\mid c; w) \;.
\end{align}
This way, we reduce $\Theta(L)$ gradient updates to $\Theta(L/l)$ (number of batches) updates.
We then use $w_f$ to compute the PS strengths for all latter phrases that occur after all former phrases in $B_f$, which only requires batched forward passes.
As a result, all former phrases in $B_f$ have the same PS strength with a latter phrase.
Intuitively, this process does not identify a specific former phrase that has high PS strength with a specific latter phrase;
instead, it identifies a segment where some phrases could form a PS with the latter phrase.
We will check in Section~\ref{sec:setup-ps-detect} if the computed PS strengths are close to the ground-truth PS strengths when we train the LM on each former phrase separately.
\subsection{Ablating the Pre-training Data}
\label{sec:ablation}

Now that we have scored a set of potential parallel structures, we conduct ablation studies to measure their effect on models' ICL ability. Specifically, we ablate PSs in pre-training data through noising, train LMs on ablated data, and compare their ICL accuracy to reference LMs trained on clean data.

Ideally, we would pre-train randomly initialized LMs from scratch on the ablated data, just as how LMs are usually pre-trained, but this is expensive.
Due to compute constraints, we follow prior work and continue pre-training off-the-shelf pre-trained LMs \cite{gururangan-etal-2020-dont, yang2022c3, ke2023continual, gupta2023continual} on clean and ablated data to study the effect of PSs on ICL.

Recall that our detection algorithm returns pairs of a former phrase and a latter phrase, as well as their PS strength.
We set a threshold on the PS strength and identify the top-$p\%$ highest-scoring pairs as PSs.
To ablate the identified PSs in the pre-training data, we replace the last token of each latter phrase with a token sampled uniformly at random
from the vocabulary.
The introduced noise allows the LM to unlearn parallel structures (and the induced ICL ability) learned earlier during pre-training from scratch.
Thus, it is more aggressive than excluding updates on tokens in parallel structures during continue pre-training, which would retain any existing ICL ability of the LM.

\section{Experiment Setup}
\label{sec:experiments}
We present the setup for continual pre-training in Section~\ref{sec:pre-training setup} and the setup for parallel structure detection in Section~\ref{sec:setup-ps-detect}. 

\subsection{Continual Pre-training}
\label{sec:pre-training setup}

\begin{table*}[t]
\fontsize{9}{9}\selectfont
\setlength{\tabcolsep}{3pt}
\centering
\begin{tabular}{p{0.1in} p{1.2in} p{3.2in} p{1.4in}}
\toprule
& Task & Description & Example \\
\midrule
\multirow{4}{*}{\rotatebox[origin=c]{90}{Natural Lang.}} & Verb Inflection & Convert a verb between present tense/past tense/past participle & ``fly'' $\Leftrightarrow$ ``flew'' $\Leftrightarrow$ ``flown'' \\
\cmidrule[0.3pt]{2-4}
& Adjective $\Leftrightarrow$ Noun & Convert an adjective to a noun or a noun to an adjective & ``exciting'' $\Leftrightarrow$ ``excitement'' \\
\cmidrule[0.3pt]{2-4}
& Case Change & Switch a word's case between lower and upper & ``hello'' $\Leftrightarrow$ ``Hello'' \\
\cmidrule[0.3pt]{2-4}
& Synonym/Antonym Clf  & Classify whether two words are synonyms or antonyms  & ``happy cheerful'' $\Rightarrow$ [$syn$] \\
\midrule\midrule
\multirow{5}{*}{\rotatebox[origin=c]{90}{Symbolic}} & Copy & Copy the input & ``hi apple'' $\Rightarrow$ ``hi apple''  \\
\cmidrule[0.3pt]{2-4}
& Last Token & Copy the last token of the input & ``hi bad orange'' $\Rightarrow$ ``orange''  \\
\cmidrule[0.3pt]{2-4}
& Search Clf & Given a token sequence $x$ and token $y$, classify if $y$ appears in $x$ & ``hi good $[del]$ hi'' $\Rightarrow$ $[yes]$ \\
\cmidrule[0.3pt]{2-4}
& Palindrome Clf & Classify if the input is a palindrome & ``apple hi apple'' $\Rightarrow$ $[yes]$\\
\cmidrule[0.3pt]{2-4}
& Pattern Completion & Complete the last token of a pattern ($a a$, $a b a$, $a b a b$ or $a a b a$) & $aba$: ``hi good'' $\Rightarrow$ ``hi'' \\
\bottomrule
\end{tabular}
\caption{\label{tab:icl-tasks}
ICL tasks. We evaluate the ICL ability of LMs on four natural language tasks and five symbolic tasks.} 
\end{table*}
\paragraph{Models}
We continue pre-training
GPT-2 models of different sizes \cite{radford2019language}: Small (117M parameters), Medium (345M parameters), Large (744M parameters), XLarge (1.6B parameters).
We choose GPT-2 models because autoregressive LMs from the GPT family have been shown to be highly successful in ICL \cite{brown2020language, Achiam2023GPT4TR}, and to balance compute cost and ICL capability following prior work \cite{wang-etal-2023-label, olsson2022context, shin2022effect, chan2022data}.


\paragraph{Data}
To minimize the distribution shift between the data used for pre-training from scratch and the data used for continual pre-training, 
we fine-tune GPT-2 on OpenWebText \cite{Gokaslan2019OpenWeb}, a publicly available version of WebText used by GPT-2.
We segment the data into context windows of length 1024.

\paragraph{Training}
We use batch size 128 and AdamW optimizers \cite{loshchilov2017decoupled} with learning rate 3e-4 for Small/Medium and 1e-4 for Large/XLarge. We early stop when the perplexity on the development set converges. 

\subsection{Parallel Structure Detection}
\label{sec:setup-ps-detect}
We construct latter phrases by partitioning each context window into segments of length $m$=12. We group former phrases into batches of $l$=128 (\cref{sec:detection}).
To measure parallel structure strengths, we fine-tune the pre-trained GPT2-Small model \cite{radford2019language} on former phrases with a learning rate of \mbox{$\eta$=1e-4}.
As a sanity check, we evaluate the similarity between the PS strengths calculated with and without the approximation of minibatch update on multiple former phrases,
and find them to strongly correlate (Pearson correlation
$+0.71$) on 10K randomly sampled context windows.
This indicates that PS strengths are relatively robust under the proposed approximations.

To evaluate LMs pre-trained on different noise rates, we ablate pre-training data with p\%=5\%, 10\%, 15\%, 20\%,
continue pre-training a LM on each, and measure their average ICL accuracy over all tasks.

\section{ICL Evaluation}
\label{sec:icl-evaluation}
\paragraph{Tasks}
We evaluate the ICL capability of LMs on four natural language tasks and five symbolic reasoning tasks (Table~\ref{tab:icl-tasks}). Natural language tasks test linguistic knowledge, while symbolic tasks test abstract reasoning that doesn't depend on the semantic meanings of tokens. 

\paragraph{Data Generation}
For natural language tasks, we prompt GPT-4 to generate the evaluation data. 
We manually check a random subset of 100 examples for each task and find no error.
For symbolic tasks, we generate the data following the procedures in \citet{li2021quantifying}.
We generate 1200 examples for each natural language task on average, and 4000 examples for each symbolic reasoning task.
We construct the in-context prompts by concatenating input-output pairs,
with delimiters between the input and the output and 
and between examples. 

\paragraph{Metric}
We evaluate models given various numbers of in-context examples (64, 96, 128), and report the average ICL accuracy as
how much the LM outperforms the random baseline (absolute).

\begin{table}[t]
\fontsize{8.2}{8.2}\selectfont
\setlength{\tabcolsep}{1.5pt}
\renewcommand{\arraystretch}{1.2}
\centering
\begin{tabular}{llcccccccccc}
\toprule
 M & Data & VrbI & A-N & Case & Syn & Cpy & LstT & Paln & Srch & Pttn & Avg \\
\midrule
\multirow{6}{*}{\rotatebox[origin=l]{90}{\textsc{GPT2-S}}} & 
\textsc{Clean} & 28.0 & 10.4 & 56.6 & 12.6 & 18.5 & 22.9 & 6.9 & 16.0 & 29.6 & 22.4 \\
& -\textsc{Rand} & 18.2 & 8.4 & 37.5 & 11.6 & 9.3 & 16.6 & 7.1 & 19.3 & 27.2 & 17.3 \\
& -PS & \textbf{3.4} & \textbf{2.6} & \textbf{17.6} & \textbf{5.2} & \textbf{0.4} & \textbf{1.1} & \textbf{-0.1} & \textbf{10.4} & \textbf{4.9} & \textbf{5.1} \\
& -Dp+PS & 8.6 & 5.6 & 29.3 & 8.4 & 2.7 & 6.3 & 7.9 & 20.8 & 20.8 & 12.3 \\
& -PS+Rp & 6.7 & 4.0 & 20.0 & 6.5 & 0.4 & 1.1 & 1.9 & 13.2 & 11.5 & 7.3 \\
\midrule
\multirow{6}{*}{\rotatebox[origin=l]{90}{\textsc{GPT2-M}}}
& \textsc{Clean} & 55.7 & 27.2 & 77.5 & 17.2 & 29.6 & 31.9 & 14.8 & 22.1 & 37.4 & 34.8 \\
& -\textsc{Rand} & 55.4 & 25.7 & 68.0 & 16.1 & 24.8 & 27.5 & 22.9 & 28.8 & 45.0 & 34.9 \\
& -PS & \textbf{28.2} & \textbf{12.0} & \textbf{52.8} & \textbf{9.3} & 0.9 & \textbf{4.7} & \textbf{11.3} & \textbf{17.6} & \textbf{14.0} & \textbf{16.7} \\
& -Dp+PS & 47.1 & 22.0 & 62.0 & 13.5 & 3.9 & 15.8 & 25.4 & 30.0 & 32.9 & 28.1 \\
& -PS+Rp & 38.4 & 16.9 & 54.8 & 10.9 & \textbf{0.6} & 6.5 & 16.7 & 23.0 & 19.3 & 20.8 \\
\midrule
\multirow{6}{*}{\rotatebox[origin=l]{90}{\textsc{GPT2-L}}}
& \textsc{Clean} & 51.1 & 33.3 & 84.5 & 21.2 & 41.0 & 38.0 & 14.5 & \textbf{17.5} & 46.3 & 38.6 \\
& -\textsc{Rand} & 60.4 & 31.7 & 75.9 & 20.6 & 46.6 & 40.7 & 23.3 & 27.8 & 56.5 & 42.6 \\
& -PS & \textbf{29.5} & \textbf{19.6} & \textbf{59.3} & \textbf{12.6} & \textbf{13.1} & \textbf{15.9} & \textbf{12.9} & 22.8 & \textbf{33.3} & \textbf{24.3} \\
& -Dp+PS & 53.3 & 27.8 & 68.6 & 17.3 & 31.0 & 31.3 & 25.1 & 31.5 & 52.8 & 37.6 \\
& -PS+Rp & 42.2 & 24.3 & 63.0 & 15.2 & 13.1 & 17.3 & 16.8 & 26.2 & 39.6 & 28.6 \\
\midrule
\multirow{5}{*}{\rotatebox[origin=l]{90}{\textsc{GPT2-XL}}}
& \textsc{Clean} & 59.2 & 35.9 & 85.3 & 30.5 & 29.4 & 37.1 & 11.9 & 17.4 & 41.6 & 38.7 \\
& -\textsc{Rand} & 61.3 & 35.9 & 77.9 & 30.2 & 30.4 & 40.8 & 17.0 & 22.3 & 54.5 & 41.2 \\
& -PS & \textbf{44.2} & \textbf{27.8} & \textbf{63.1} & \textbf{19.1} & \textbf{5.5} & \textbf{10.0} & \textbf{5.6} & \textbf{12.9} & \textbf{27.9} & \textbf{24.0} \\
& -Dp+PS & 62.4 & 35.6 & 73.5 & 25.2 & 23.5 & 27.9 & 14.6 & 21.6 & 54.2 & 37.6 \\
& -PS+Rp & 59.8 & 33.2 & 67.4 & 22.6 & 10.6 & 17.7 & 11.3 & 18.2 & 45.7 & 31.8\\
\bottomrule
\end{tabular}
\caption{\label{tab:ablation} We measure the effect of different data ablations on the ICL ability of pre-trained LMs. Results show that parallel structures are crucial for LMs to acquire ICL. Pre-training on data with parallel structures ablated consistently incurs a larger drop in ICL accuracy compared to pre-training on data with random tokens ablated ({51.1\%} vs {1.5\%} relative drop in accuracy averaged across model sizes). We also compare parallel structures to n-gram repetitions (Rp) and long-range dependency (Dp) and find parallel structures to have larger effect on ICL. The pre-training setting that incurs the largest drop in ICL performance is bold for each task and model size.}
\end{table}

\section{Results}
\label{sec:results}
We first measure the effect of parallel structures on ICL (\mbox{Section~\ref{sec:results-ps-effect}}), then compare their effect to other structures identified by prior work (\cref{sec:results-compare}), and finally analyze characteristics of parallel structures in the pre-training data (\cref{sec:analysis}).


\subsection{Measuring the Effect of Parallel Structures on ICL}
\label{sec:results-ps-effect}

To measure the effect of parallel structures on ICL, we continue pre-training the LM on ablated data ($-$\textsc{PS}), and compare its ICL accuracy with LMs continually pre-trained on the clean data (\textsc{Clean})
and the randomly noised data ($-$\textsc{Rand}),
where tokens sampled uniformly at random from the clean data are ablated.
We ablate the same amount of tokens in $-$\textsc{PS} and $-$\textsc{Rand}.
%

\paragraph{Ablating parallel structures hurts ICL.}
In \mbox{\cref{tab:ablation}},
both $-$\textsc{Rand} and $-$\textsc{PS} hurt ICL performance compared to
\textsc{Clean}, which is expected as data noise can hurt model performance in general. However, ablating PSs is particularly detrimental to ICL performance compared to ablating random tokens of the same amount ({51.1\%} vs {1.5\%} relative drop in accuracy averaged across model sizes).

\begin{table}[t]
\fontsize{9}{9}\selectfont
\renewcommand{\arraystretch}{0.4}
\centering
\begin{tabular}{lcc}
\toprule
& -\textsc{Random} & -PS \\
\midrule[0.3pt]
GPT2-S & 50.4 & \textbf{49.9} \\
\midrule[0.3pt]
GPT2-M & 54.0 & \textbf{53.9} \\
\midrule[0.3pt]
GPT2-L & 60.0 & \textbf{59.6} \\
\midrule[0.3pt]
GPT2-XL & \textbf{62.1} & 62.4 \\
\bottomrule
\end{tabular}
\caption{\label{tab:finetuning} {Fine-tuning} accuracy of LMs. Contrary to the ICL results, LMs further pre-trained on data with parallel structures ablated have comparable fine-tuning accuracy as LMs trained on randomly ablated data.}
\end{table}

\paragraph{Ablating PSs does not hurt task ability.}
One caveat in the above numbers is that ICL accuracy confounds ICL ability with task ability.
Low ICL accuracy can be caused by a failure to identify the task based on ICL examples (ICL ability) or by a failure to perform the identified task (task ability).
To disentangle the two sources of failure, we evaluate a LM's task ability by measuring its \textit{fine-tuning} accuracy.
Specifically, for each task we fine-tune the LM on 128 examples and report the average task accuracy.
Contrary to the ICL results where ablating parallel structures ($-$\textsc{PS}) consistently leads to larger accuracy reduction than ablating random tokens ($-$\textsc{Rand}), the two ablations have comparable fine-tuning accuracy as shown  in Table~\ref{tab:finetuning}. 
Thus, the drop in ICL accuracy from ablating parallel structures is mainly due to a drop in ICL ability, not task ability. 

\subsection{Comparing Parallel Structures with Other Structures}
\label{sec:results-compare}
We compare parallel structures with two other structures of pre-training data hypothesized to produce ICL: n-gram repetitions and long-range dependency (Table~\ref{tab:ablation}).

\paragraph{Parallel structures that are not n-gram repetitions are also important for ICL.}
Prior work has shown that ICL is closely related to n-gram repetitions in the pre-training data \cite{yan2023understanding, olsson2022context}.
N-gram repetitions are a subcategory of parallel structures where the former and latter phrases are identical.
Are parallel structures crucial for ICL only because they include n-gram repetitions?
To answer this question, we measure the effect of parallel structures that are not n-gram repetitions on ICL, denoted as $-\textsc{PS} + \textsc{Rp}$.
Specifically, during PS scoring we exclude phrase pairs that end with the same bigram, e.g., ``mortality rates in 2013'' and ``mortality rates again in 2013''. We then take the top-$p\%$ PSs and perform ablation as described in \cref{sec:ablation}. 

Pre-training on $-$\textsc{PS} $+$\textsc{Rp} consistently incurs a larger drop in ICL performance compared to ablating random tokens of the same amount (37.9\% vs. 1.5\% relative reduction in accuracy averaged across model sizes), which indicates that 
parallel structures that are not n-gram repetitions are also important for LMs to acquire ICL. We conjecture that pre-training on diverse parallel structures helps LM generalize to various downstream tasks where copying alone is insufficient (e.g., synonym/antonym classification and palindrome classification).

In particular, we observe that ablating parallel structures that are not repetitions incurs a large drop in ICL accuracy on the copy task as well (81.8\% relative reduction in accuracy averaged across model sizes), even though all parallel structures that are repetitions are preserved. This indicates that LMs learn to generalize between parallel structures/in-context examples of different tasks.


\paragraph{Parallel structures have a larger effect on ICL than long-range dependency.} 
Prior work identified long-range dependency in pre-training data as crucial for LMs to acquire ICL \cite{shi2023context}.
Parallel structures are a subcategory of long-range dependency, where the dependency is the similarity between two phrases from the same distribution.
Are PSs crucial for ICL only because they capture long-range dependency?
In other words, is long-range dependency that are not PSs equally crucial for ICL?
To answer this question, we measure the effect of long-range dependency that is not parallel structures on ICL, denoted as $-\textsc{Dp}+\textsc{PS}$.
Motivated by \citet{sun-etal-2021-long, olsson2022context},
for each latter phrase $(c_l, x_l)$ in a segment $b$
whose context length is at most $m$,
it has long range dependency if including additional context improves the log probability of $x_l$ under the language model.

Specifically, the long context includes all previous tokens in the context window $a$ as illustrated below:
\vspace{-1.2em}
\\[0.5ex]
\begin{center}
\begin{tikzpicture}[scale=0.9]
     \fill[red!20] (0,0) rectangle (8,0.5);
     \node[anchor=west] (a) at (0,0.25) {$\color{red}{a}$};

     \fill[green!20] (4,0) rectangle ++(2,0.5);
     \node[anchor=west] (b) at (4,0.25) {$\color{green!50!black}{b}$};

     \node[anchor=west] (x) at (4+1.25,0.25) {$x_l$};
     \draw[-latex, thick, <->] (x.north) to[out=180-5,in=5] node[above,               midway]{$p(x_l\mid \text{long context})$} (a.north);
     \draw[-latex, thick, <->] (x.south) to[out=180+10,in=-10] node[below,              midway]{$p(x_l\mid \text{short context})$} (b.south);
 \end{tikzpicture} \\[0.5ex]
\end{center} 
Formally, given a context window $a$, for each $(c_l, x_l)$ where $x_l=a_i$,
we measure the long-range dependency strength of the phrase by
\begin{align}
    &\beta(c_l, x_l=a_i) \\
    = & \log p(a_i \mid a_{<i}; w) 
    - \log p(a_i \mid c_l; w)
\end{align}
Same as detecting parallel structures, we use pre-trained GPT2-Small as the language model for scoring 
and ablate the top-$p\%$ $(c_l, x_l)$ with long range dependency by replacing $x_l$ with a random token.

Pre-training on $-$\textsc{Dp} $+$\textsc{PS} consistently incurs a smaller drop in ICL performance compared to pre-training on $-$\textsc{PS} on all four model sizes
({17.5\%} vs {51.1\%} relative reduction in accuracy averaged across model sizes).
This indicates that parallel structures are crucial for ICL not because they capture long-range dependency, and that
parallel structures have a larger effect on ICL than long-range dependency.

\subsection{Analyzing Characteristics of Parallel Structures}
\label{sec:analysis}
In addition to the ablation results, we analyze characteristics of the detected parallel structures in pre-training data, and find that they are suggestive of ICL abilities we observe on large LMs.
These links between parallel structures and ICL present additional evidence that PSs produce ICL, and more importantly, open up new directions/methods to study ICL by tracing back to PSs in the pre-training data.

\paragraph{Parallel structures exhibit diverse patterns.} 
We find that the detected parallel structures in pre-training data exhibit diverse patterns, including
n-gram repetitions, synonyms, text formats (e.g., ``\textbackslash{}n\textbackslash{}n'' followed by a $\langle$year number$\rangle$),
syntactic constituents (e.g., a $\langle$pronoun$\rangle$ followed by a $\langle$verb$\rangle$), punctuation and line break patterns, and more complicated ones that require reasoning and knowledge (e.g., a $\langle$basketball player$\rangle$ followed by $\langle$their position$\rangle$).
Each type of parallel structures corresponds to the text distribution of some ``task'' that the model needs to learn in-context, so the types of parallel structures in the pre-training data corresponds to the number of pre-training ``tasks''. 
Prior work also hypothesized the importance of task diversity for learning new linear regression tasks 
\cite{raventos2023pretraining} and the importance of domain diversity for ICL \cite{shin2022effect}. Our work detects the in-context ``tasks'' in real pre-training data, and finds that their diversity is crucial for LMs to acquire ICL.


\paragraph{Parallel structures span long distances.} 
We measure the distance (i.e.\ number of tokens) between the former and latter phrases in the identified PSs, and find that parallel structures often span long distances (skewed to the right with an average of 343 tokens, a median of 292 tokens, and a standard deviation of 275 tokens).
Pre-training on parallel structures spanning long distances may encourage LMs to use patterns of early tokens in the context to predict the next token.
This ability may explain why LMs do not forget early examples in in-context prompts \cite{li-qiu-2023-finding} and achieve monotonically higher accuracy with more ICL examples on most tasks \cite{brown2020language}.


\section{Related Work}


\paragraph{Effect of Pre-training Data on ICL.} Prior work has studied what structures of pre-training data are crucial for LMs to acquire ICL.
We introduce them below and discuss their relations to parallel structures. 

\emph{Long-range dependency.}
One line of work showed that pre-training LMs on data with long-range coherence produces ICL.
\citet{xie2021explanation} generated a synthetic dataset where 
each context window consists of multiple segments sampled from the same Hidden Markov Model, and showed that pre-training on this synthetic dataset produces ICL. \citet{shi2023context} verified the importance of long-range coherence on natural language text by empirically showing that concatenating relevant text during pre-training improves ICL. Parallel structures are a special kind of long-range dependency that is more important for ICL.


\textit{N-gram repetitions.}
\citet{olsson2022context} found that n-gram repetitions are closely related to ICL through induction heads: LMs learn induction heads from n-gram repetitions, and this process happens concurrently with the emergence of ICL during pre-training. \citet{yan2023understanding} claimed that LMs learn token co-occurrence reinforcement from n-gram repetitions, which is essential for ICL. 
Parallel structures include n-gram repetitions as a subcategory, but also include less structured patterns that are also crucial for ICL.

\emph{Diversity.} \citet{shin2022effect} found that increasing corpus diversity by merging datasets of different domains improves ICL.
Our results show that diverse parallel structures are crucial for ICL.

\textit{Long-tail tokens.}
\citet{han-etal-2023-understanding} identified supportive pre-training data with similar gradients as in-context examples, and found that the supportive data has higher density of long-tail tokens compared to natural text. Instead of studying the effect of pre-training data on ICL, \citet{chan2022data} studied the effect of in-context tuning (i.e. training on in-context prompts \cite{chen2022meta})
data on ICL, and also found that increasing the number of long-tail classes improves ICL.
It is unclear how long-tail tokens are related to parallel structures.


\paragraph{Mechanistic Interpretability of ICL.}

Prior work has proposed different theories to explain how ICL works. We introduce them below and discuss the connection between those mechanisms and parallel structures.

\textit{Induction heads.} \citet{olsson2022context} claimed that LMs perform ICL via induction heads: attention heads that attend to a previous occurrence of a similar phrase and copy from it. Their work supported their claim by showing that ICL and induction heads appear concurrently during pre-training. As a follow-up work, \citet{wang-etal-2023-label} studied how LMs use attention heads to perform ICL, and found that label words of ICL examples aggregate information processed in shallow layers and provide anchors for induction heads.
We conjecture that LMs may also use induction heads to predict parallel structures, and leave it to future work.

\textit{Implicit gradient descent.} Multiple concurrent work~\cite{akyurek2022learning, von2023transformers, mahankali2023one} claimed that LMs perform ICL via implicit gradient descent, where one layer of model inference on in-context examples corresponds to one step of gradient descent on those examples. This group of work supported its claim on linear regression tasks, which is then generalized to natural language tasks by \citet{dai-etal-2023-gpt}.
We detect parallel structures using gradient descent, and an interesting future direction is to explore if the LM's behavior on parallel structures in text also resembles gradient descent.


\section{Conclusion}
We study what structures of the pre-training data yield in-context learning, and hypothesize that parallel structures are crucial for LMs to acquire ICL ability.
We verify our hypothesis with ablation experiments on real pre-training data, where we find that ablating parallel structures incurs a significant drop in ICL performance.
Detailed analysis further reveals that parallel structures are more important than n-gram repetitions and long-range dependency for ICL, and exhibit diverse linguistic patterns. 
We hope our findings can inspire future methods to construct better pre-training data to improve ICL performance, and to better understand the source of emergent ICL ability.


\section{Limitations}
Our work has several limitations that we leave to future work. First, due to limited computational resources we only experiment with models up to 1.5 billion parameters. Future work should scale up our experiments to larger LMs and explore pre-training randomly initialized LMs from scratch. Second, despite our efforts in creating a set of diverse and representative tasks to evaluate ICL ability, most tasks are relatively straightforward due to limitations imposed by the LM size we experiment with (i.e. our experimented LMs fail on most complex tasks). Future work should study evaluate ICL ability on more complicated tasks with larger LMs. Third, our study focuses on parallel structures and ICL in the text modality. Future work should study the role of parallel structures in multi-modal ICL.
\section{Acknowledgements}
YC is supported by an Avanessians Doctoral Fellowship. CZ is supported by Shanghai Frontiers Science Center of Artificial Intelligence and Deep Learning, NYU Shanghai.

This research is supported in part by the Office of the Director of National Intelligence (ODNI), Intelligence Advanced Research Projects Activity (IARPA), via the HIATUS Program contract \#2022-22072200005. The U.S. Government is authorized to reproduce and distribute reprints for governmental purposes notwithstanding any copyright annotation therein. This work was funded in part by the US Department of Defense under the DARPA CCU program. Any opinions expressed herein are those of the authors and do not necessarily reflect the views of the U.S. Department of Defense, ODNI, IARPA, or any other agency of the U.S. Government.

\bibliography{custom}
\bibliographystyle{acl_natbib}

\appendix



\end{document}